# Hesitation is defeat? Connecting Linguistic and Predictive Uncertainty

Gianluca Manzo[1]
Julia Ive[2]
Queen Mary University of London
University College London

*Abstract*—Automating chest radiograph interpretation using Deep Learning (DL) models has the potential to significantly improve clinical workflows, decision-making, and large-scale health screening. However, in medical settings, merely optimising predictive performance is insufficient, as the quantification of uncertainty is equally crucial. This paper investigates the relationship between predictive uncertainty, derived from Bayesian Deep Learning approximations, and human/linguistic uncertainty, as estimated from free-text radiology reports labelled by rule-based labellers. Utilising BERT as the model of choice, this study evaluates different binarisation methods for uncertainty labels and explores the efficacy of Monte Carlo Dropout and Deep Ensembles in estimating predictive uncertainty. The results demonstrate good model performance, but also a modest correlation between predictive and linguistic uncertainty, highlighting the challenges in aligning machine uncertainty with human interpretation nuances. Our findings suggest that while Bayesian approximations provide valuable uncertainty estimates, further refinement is necessary to fully capture and utilise the subtleties of human uncertainty in clinical applications.

## I. Introduction

Chest radiography, the most widely used imaging test worldwide, plays a crucial role in screening, diagnosing, and managing numerous life-threatening diseases. Automating the interpretation of chest radiographs to match the expertise of practising radiologists could significantly enhance various medical settings. This includes improving workflow prioritisation, supporting clinical decisions, and facilitating large-scale screening and global population health initiatives.

Recent advances have enabled Deep Learning (DL) models to achieve performance comparable to clinical experts. However, optimising predictive performance alone is not sufficient in the healthcare domain. Given the high-stakes consequences of clinical decision-making, particularly when positive cases are missed (false negatives), it is essential for clinicians to receive trustworthy predictions. This trust can be established through models that not only provide point estimates (which can fluctuate significantly) but also quantify the uncertainty of such estimates (predictive uncertainty). For this reason, I am employing Bayesian Deep Learning approximations to quantify predictive uncertainty in detecting chest complications.

For this analysis, I am using as input the free-text reports that are produced by radiologists upon assessment of patient chest X-rays. This represents a Natural Language Processing (NLP) problem that remains understudied in the healthcare domain. Due to uncertainties inherent in radiograph interpretation, evaluation of the reports by several doctors can lead to differing opinions (human uncertainty). Furthermore, rule-based labellers, often used as ground truth due to the lack of human-annotated data, can also disagree on outcomes (linguistic uncertainty). By employing (approximations of) Bayesian methods to produce uncertainty estimates for prediction models, I aim to investigate the correlation between predictive uncertainty and human/linguistic uncertainty in binary classification.

This paper will first review previous work on the problem (Section II). Subsequently, I will present an overview of methodologies (Section III) used for the data/outcome of interest, prediction modelling and uncertainty estimation, followed by the respective results (Section IV). Finally, I will conclude my analysis by discussing the results (Section V) and suggesting how current work may be extended (Section VI).

## II. Related Work

### A. Automating Chest Radiograph Interpretation

Deep Learning has replaced traditional Machine Learning in many medical applications (Purushotham et al., 2018) due to its flexibility, scalability, minimal need for feature engineering, and superior performance on larger datasets. Remarkably, DL models have even achieved similar performance to human experts (Topol, 2019).

In the field of radiology, the release of large, labelled datasets of high quality has the potential to advance automated chest radiograph interpretation. The ChestX-ray14 dataset (Wang et al., 2017) was, until recently, the most used benchmark for the development of chest radiograph interpretation models. The subsequent co-release of two additional large datasets, CheXpert (Irvin et al., 2019) and MIMIC-CXR (Johnson et al., 2019), has further increased interest in the field. The CheXpert Competition (Irvin et al., 2019) allowed significant progress to be made in classification performance, with several Deep Learning models/architectures reaching an AUC of 92%-93% (e.g., Pham et al., 2021 and Yuan et al., 2021). The goal of the competition was to predict 14 common chest radiographic observations from multi-view chest radiographs. Of these 14 observations, CheXpert identified 5 observations of high clinical importance and prevalence: Atelectasis, Cardiomegaly, Consolidation, Edema, and Pleural Effusion.

### B. Uncertainty-Aware Deep Learning

While optimising classification performance is important, uncertainty provides more insight than confidence into what a model does not know (Ovadia et al., 2019). For this reason, a range of uncertainty-aware models have been increasingly explored in clinical NLP, particularly in radiology. Liu et al. (2022) found that Gaussian Processes (Rasmussen and Williams, 2006) provide superior performance in quantifying the risks of uncertainty labels in radiology reports. The models are optimised for higher predictive performance (accuracy) and lower negative log predictive probability (NLPP), which penalises both over-confident incorrect predictions and under-confident correct predictions. In contrast, my study employed BERT and was optimised for higher accuracy and higher point-biserial correlation coefficient (Kornbrot, 2014:1), which will be addressed in detail later. Given the computational expense of Gaussian Processes (complexity of $O(n^3)$), two popular Bayesian Deep Learning approximations were explored: Monte Carlo (MC) Dropout (Gal and Ghahramani, 2016) and Deep Ensembles (Lakshminarayanan et al., 2017). Unlike Liu et al. (2022), who kept the 'uncertain' class separate in their models (3-class classification), my analysis handled the uncertainty class by binarising the label in 3 ways, which will be discussed later.

State-of-the-art bidirectional encoder representation from transformers (BERT) and its variants (BioBERT, ClinicalBERT) have been shown to be reliable tools for radiologist report evaluation (Liu et al., 2021), justifying my choice of model. While research on uncertainty-aware Deep Learning in healthcare is growing, the field is still largely unexplored, with no consensus on optimal methods for quantifying uncertainty (Loftus et al., 2022). Furthermore, for NLP of radiology reports, the quality of reporting in available studies is often suboptimal (e.g., insufficient use of common datasets to benchmark systems), hindering comparison and reproducibility. The latter, together with the explainability of models, would be needed to move applications into clinical use (Casey et al., 2021, Davidson et al., 2021).

*C. Uncertainty Detection*

Human-annotated data in clinical NLP of radiology reports are limited and expensive, leading to the use of rule-based labellers as ground truth. Together with their large dataset, Irvin et al. (2019) released the CheXpert labeller, which uses rules predefined by experts to extract, classify, and aggregate observations in a 3-stage process. The CheXpert labeller builds upon (and improves) NegBio (Peng et al., 2018), another popular rule-based labeller used to annotate previous datasets of a wider domain. Specifically, both labellers were evaluated on mention extraction, negation detection, and uncertainty detection, with CheXpert significantly outperforming NegBio across all chest observations and tasks (Irvin et al., 2019).

CheXpert is an important step towards human-like evaluation. Nonetheless, rule-based labellers can still only identify explicit uncertainty (with immediate lexical markers such as 'possibly', 'cannot be ruled out', etc.), as they often ignore uncertainty in causality, medical opinions, and so on (implicit uncertainty). Furthermore, rule-based labellers cannot distinguish between different types of uncertainty. For this reason, Turner et al. (2021) proposed to reformulate uncertainty detection as a multi-class classification problem, as they introduced a new taxonomy with 9 discrete types of uncertainty.

Even when CheXpert identifies uncertainty, it is unable to yield probabilistic outputs. Furthermore, the labeller is computationally slow and nondifferentiable, so it cannot be employed in applications that require gradients to flow through the labeller. To solve these issues, McDermott et al. (2020) proposed CheXpert++, a BERT-based, high-fidelity approximation to CheXpert that achieves around 99.8% parity with the popular rule-based labeller.

*D. Referral Learning*

Uncertainty-aware Deep Learning enables the comparison of predictive uncertainty with human uncertainty through a Referral Learning approach (Popat and Ive, 2023). In this method, models refer (delegate) their most uncertain cases to human experts for further evaluation, akin to a junior doctor referring complex cases to a specialist. This collaborative approach has the potential to outperform both the model and the human working independently. Referral Learning is particularly useful when there is availability of data annotated by multiple doctors (multiple-label scenario i.e., several labels per study), as this allows human uncertainty to be estimated/predicted as well. Nonetheless, a simpler type of Referral Learning based on a single-label scenario (one label per study) was also explored. The goal of the study was the detection of two benchmarking mental health conditions (dementia and depression) using text-based clinical data (respectively, patient descriptions of household scenes and discharge summary notes following hospital admission). Nonetheless, the Referral Learning approach generalises to any healthcare outcome (including chest radiographic observations) that has been evaluated via uncertainty-aware solutions. Popat and Ive's (2023) human-model cooperation setup allowed the best model to exceed the state-of-the-art performance by referring about 15% of cases to human experts.

### III. METHODOLOGY

*A. Data and Outcome of Interest*

For my analysis, I evaluated radiology reports from MIMIC-CXR-JPG (Jonhson et al., 2024), which is a conveniently processed version of MIMIC-CXR. The free-text reports are labelled using CheXpert and NegBio. Both labellers convert radiology reports into positive, negative, or uncertain labels (as shown in Fig. 1, Fig. 2, and Fig. 3 respectively). CheXpert performs better than NegBio in the domain of radiology interpretation and, for this reason, CheXpert was used as ground truth during model training. As preprocessing the free-text reports improved both model performance and uncertainty estimation, I explored several methods using NLTK and spaCy. Examples of preprocessing techniques I employed are stopword removal (excluding negations such as 'no' and 'not'), removing headers section titles such as 'FINAL REPORT', lowercasing and lemmatising tokens, and so on.

As in Liu et al. (2022), I chose Edema as the outcome of interest because of the large data size, the greater inconsistency in labels between the two labellers, and the more balanced split between classes. Since we do not have ground truth labels for multiple human experts, human uncertainty is approximated by looking at cases where CheXpert and NegBio disagree, as well as evaluating the two labellers separately. To accurately assess predictive uncertainty, it is imperative to first determine the appropriate methodology for integrating uncertainty labels within the model training process. For this, Irvin et al. (2019) evaluated the following approaches: Ignoring (i.e. ignoring the uncertainty labels altogether during training), Binary Mapping (i.e. replacing 'uncertain' labels with 'positive' in U-Ones and with 'negative' in U-Zeros), Self-Training (train model ignoring uncertain labels, then use the model to predict/re-label uncertain labels to positive/negative), and 3-Class Classification (uncertain labels are treated as a separate class during training). I focused on Binary Mapping, comparing U-Ones with U-Zeros on model performance (binary classification) and uncertainty estimation. Notably, I evaluated a new binary approach that randomly maps uncertain labels to positive/negative in a 50-50 split. This method, which was renamed U-Random, can be considered a middle-ground approach between U-Ones and U-Zeros.

Fig. 1.  Example of report labelled 'positive' by CheXpert and NegBio.

Fig. 2.  Example of report labelled 'negative' by CheXpert and NegBio.

Fig. 3.  Example of report labelled 'uncertain' by CheXpert and NegBio.

### B. Modelling and Uncertainty Estimation

> *As compared to the prior examination dated \_\_\_, there has been slight interval increase in **now moderate pulmonary interstitial edema** and central pulmonary vascular congestion. A background of prominent interstitial markings likely reflects underlying interstitial lung disease,*

> *A right subclavian central line remains in position with its tip in the distal SVC near the cavoatrial junction. Dual-lead left-sided pacer with its leads terminating over the expected location of the right atrium and right ventricle, respectively. Markedly low lung volumes with crowding of the pulmonary vasculature with indistinct vasculature on the left **raising a concern for asymmetric pulmonary edema**. Patchy opacity at the right base may reflect atelectasis, although pneumonia cannot be entirely excluded. A mid to distal left clavicular fracture is again seen. No pneumothorax.*

My model of choice is the popular BERT, state-of-the-art for many NLP tasks (Devlin et al., 2018). Developed by Google, BERT leverages the transformer-based architecture to understand the context of words in a sentence by analysing both preceding and succeeding text (bidirectional context), enabling more nuanced and accurate language comprehension. I trained BERT on the free-text radiology reports via the recommended train/validation/test set splits from MIMIC-CXR-JPG. Specifically, I implemented BERT as a binary classifier via single run (train on the training set, compare on the validation set) and K-fold cross-validation (with K=5). The test set was used to evaluate the final model. I also used BERT's built-in tokeniser to tokenise the reports. The loss function of choice is cross entropy, which was minimized via the AdamW optimiser. AdamW, an adaptive learning rate optimisation algorithm, decouples weight decay from the gradient update by applying it directly to the weights post-gradient update (Loshchilov and Hutter, 2019). My performance metrics of choice are accuracy and F1 score. Hyperparameters were found arbitrarily (learning rate = $2 \cdot 10^{-5}$, weight decay = $10^{-4}$, epsilon = $10^{-7}$, batch size = 128, report max length = 512 (#)). As a larger number of epochs did not significantly improve performance, I limited the number of epochs to 3. Last, but not least, my analysis was implemented using the PyTorch framework (with its Dataset and DataLoader primitives).

Quantifying predictive uncertainty is crucial in Deep Learning models, as they only generate point estimate predictions that can vary significantly due to factors like random seed changes. Predictive uncertainty can be decomposed into epistemic and aleatoric uncertainty (Hüllermeier and Waegeman, 2021). Epistemic uncertainty results in a wider predictive distribution, increasing the likelihood of including the decision boundary. This type of uncertainty, which can theoretically be reduced with additional information, represents the reducible part of the total uncertainty. In contrast, aleatoric uncertainty arises from inherent noise in the data, reflecting irreducible variability. Overall, predictive uncertainty results in probabilities close to the decision threshold (0.5 for this binary classification task), indicating a lack of confidence in the prediction.

To quantify uncertainty, I implemented approximations of Bayesian Deep Learning, which extends Deep Learning by incorporating probability distributions into model weights (Blundell et al., 2015). I explored 2 methods: MC Dropout and Deep Ensembles. MC Dropout originates from dropout, a regularisation method used to prevent overfitting in Deep Learning. During training, a fraction of the neurons' outputs (dropout rate) is set to zero to prevent a single neuron from becoming overly dependent on neurons in the previous layer. At inference, the prediction is deterministic. In contrast, MC Dropout implements dropout during both training and testing, resulting in multiple (random) predictions per study due to varying dropout configurations at test time. Deep Ensembles also generate multiple predictions per datum, although via a different approach. As a form of Ensemble Learning, Deep Ensembles combine the predictions of several machine learning models. Unlike MC Dropout, which creates multiple predictions from a single model, Deep Ensembles involve training several independent Deep Neural Networks, each with a different random seed, producing one prediction per model.

As MC Dropout and Deep Ensembles both produce multiple predictions per study, it is possible to aggregate them via measures of central tendency and dispersion. Specifically, the average of predictions was used as the final estimate of probability (with decision boundary at 0.5). I also computed (sample) standard deviation of predictions (Predictive Standard Deviation, or PSD for short) as an estimate of epistemic uncertainty. I used predictive standard deviation instead of variance as the former provides a more intuitive measure of dispersion and uncertainty since it is in the same units as the predicted values. Last, but not least, mean predictions were used to compute Predictive Entropy (PE) (Shannon, 1948), which for binary classification is defined as:

$$\text{PE} = \text{H}(\hat{p}) = -\hat{p}\log(\hat{p}) - (1-\hat{p})\log(1-\hat{p}) \qquad (1)$$

Equation (1) defines $\hat{p}$ as the average of predictions (generated by different dropout configurations as in MC Dropout or by different models as in Deep Ensembles) for a given study. PE will be used as an estimate of predictive (total) uncertainty. It is highest at $\hat{p} = 0.5$ (decision boundary) and lowest when $\hat{p}$ is 0 or 1 (maximum confidence for negative/positive class).

As we do not have ground truth labels from human experts, human uncertainty was simulated via CheXpert/NegBio in two ways:

- Ground truth labels coming from both labellers were first binarised using U-Random, U-Ones, or U-Zeros. Subsequently, If CheXpert and NegBio labels disagree, then there is 'human' (linguistic) uncertainty (TLD = 1), otherwise the labellers are certain (TLD = 0).

- What about evaluating CheXpert and NegBio separately? In this case, I retrieved the original, 3-class labels (positive, negative, uncertain). If CheXpert labels a study as 'uncertain', then there is human uncertainty (Chex Uncertain = 1), whereas if a study is labelled positive or negative then the labeller is certain (Chex Uncertain = 0). The same rationale applies to NegBio (Neg Uncertain). It is worth noting that, as all training was done using CheXpert as ground truth, NegBio labels are considered out-of-distribution (OOD) labels.

Studies with true label disagreement (or uncertain CheXpert/NegBio labels) would ideally have higher PE (predictive uncertainty) and PSD (epistemic uncertainty), and vice versa. To compare a continuous variable Y (PE or PSD) with a binary variable X (TLD, Chex Uncertain, or Neg Uncertain), I used the Point-Biserial Correlation Coefficient ([Kornbrot, 2014:1](#)), defined as the following:

$$R_{pb} = \frac{\bar{Y}_1 - \bar{Y}_0}{S_Y} \cdot \sqrt{\frac{N_1 \cdot N_0}{N \cdot (N-1)}} \qquad (2)$$

In (2), $S_Y$ is the sample standard deviation of the continuous variable Y. Furthermore, $Y_1$ is the mean value of Y for all uncertain studies (X = 1), whereas $Y_0$ is the mean value of Y for all certain studies (X = 0). Lastly, $N_1$ is the sample size of the uncertain group, $N_0$ is the sample size of the certain group, and N is the overall sample size. Just like the standard Pearson Correlation, values for the Point-Biserial Correlation always fall within the [−1, +1] range, where +1 indicates perfect positive correlation, −1 indicates perfect negative correlation, and 0 indicates no association at all. Additionally, the square of this correlation quantifies the effect size, as it represents the proportion of variability accounted for by the relationship between X and Y. Ideally, we aim for $R_{pb}$ to be positive and as high as possible, meaning that when there is 'human' (linguistic) uncertainty, our Bayesian approximations capture this via high predictive/epistemic uncertainty.

## IV. RESULTS

### A. Summary Statistics

MIMIC-CXR-JPG provides 227,835 radiographic studies, of which 227,827 were labelled by CheXpert/NegBio (8 reports could not be labelled due to lack of findings/impression section). Each report is labelled for the presence of 14 observations, as 65,833 reports were labelled for the presence of Edema. Table 1 shows the recommended training/validation/test split (respectively, 97.2%/0.8%/1.9% of the whole dataset). Furthermore, Table 2 and Table 3 show the distribution of labels according to CheXpert and NegBio respectively. It is worth noting that, for both CheXpert and NegBio, while the train and validation sets roughly have a 40%/40%/20% label split (positive/negative/uncertain), the test set has a higher proportion of positive cases (around 52%-53%) as well as a lower proportion of negative cases (around 24%-26%). Furthermore, CheXpert is more conservative than NegBio, as it includes around 5%-7% more uncertain labels across all sets (train/validation/test).

CheXpert and NegBio can disagree on outcomes, as shown in Fig. 4. Specifically, around 3.6% of studies have true-label disagreement (TLD). While the recommended train set sticks to the same proportion, the validation set has a lower percentage of TLD (around 2.7%) whereas the test set has a higher percentage (around 4.2%). All in all, there are imbalances across sets that may suggest employing stratified sampling. Nonetheless, I adhered to the splits as they were recommended by MIMIC-CXR-JPG.

TABLE 1
MIMIC-CXR-JPG DATASET OVERVIEW

| Labelled Studies | Labelled for Edema | Train | Validation | Test |
|---|---|---|---|---|
| 227,827 | 65,833 | 64,003 | 553 | 1,277 |

TABLE 2
CHEXPERT LABEL DISTRIBUTION

| Label | Train | Val | Test |
|---|---|---|---|
| Positive | 26,093 (40.8%) | 242 (43.8%) | 683 (53.5%) |
| Negative | 25,133 (39.3%) | 203 (36.7%) | 305 (23.9%) |
| Uncertain | 12,777 (20%) | 108 (19.5%) | 289 (22.6%) |

TABLE 3
NEGBIO LABEL DISTRIBUTION

| Label | Train | Val | Test |
|---|---|---|---|
| Positive | 26,105 (40.8%) | 238 (43.0%) | 672 (52.6%) |
| Negative | 25,699 (40.1%) | 214 (38.7%) | 330 (25.8%) |
| Uncertain | 12,199 (19.1%) | 101 (18.3%) | 275 (21.5%) |

> *The heterogeneous, perihilar abnormality that developed in both lungs between __ and __ worsened subsequently, has improved over the past 24 hours. The distribution suggests atypical pneumonia **rather than edema, but edema is not excluded**. Heart is normal size, though increased slightly compared to __. Pleural effusions are small if any. No pneumothorax.*

Fig. 4. Example of report labelled 'uncertain' by CheXpert but 'negative' by NegBio.

### B. Model Performance (U-Random)

I implemented BERT on the U-Random labels using the hyperparameters outlined in (#). Specifically, I trained the model on the training set and then evaluated it on the validation set (single run). It achieved validation accuracy of around 87.9% (Table 4). Its (validation) F1 score is 88.4%, with 91% precision and 85.8% recall. While the recall shows that the model was effective at identifying most true positive cases, additional analysis would be needed to increase recall further, especially in a medical setting where missing a positive case could have serious or even life-threatening consequences. For a more systematic way to validate the model, I also performed 5-fold cross-validation, which returned similar performance results (i.e. around 88%-89% for both accuracy and F1 score).

In theory, Bayesian Deep Learning models and approximations should outperform standard Deep Learning models because their predictive distributions offer a more accurate assessment of which class to predict. For example, suppose that a given study has a predictive distribution that is uniform on [0.25, 0.65]. In this case, the BDL model can use the predictive mean of 0.45 to make the prediction. In contrast, a DL model might incorrectly classify the instance based on a point estimate within (0.5, 0.65]. Practically, we find that MC Dropout and Deep Ensembles do not significantly exceed BERT's cross-validation performance (in fact, in some cases, they perform slightly worse than the underlying model). This may be due to MC Dropout and Deep Ensembles both generating only 10 predictions per study due to computational constraints. The cross-validation performance results for both MC Dropout and Deep Ensembles are all in the 88%-89% region (Table 4), with MC Dropout narrowly beating Deep Ensembles on both accuracy and F1 score.

TABLE 4
U-RANDOM MODEL PERFORMANCE

| Performance Metric | Method | | | |
|---|---|---|---|---|
| | *Single Run* | *Cross Validation* | *MC Dropout* | *Deep Ensembles* |
| Accuracy | 0.8788 | 0.8863 | 0.8865 | 0.8851 |
| F1 Score | 0.8835 | 0.8819 | 0.8844 | 0.8831 |
| Precision | 0.9104 | | | |
| Recall | 0.8581 | | | |

*C. Predictive Uncertainty versus Human Uncertainty*

Table 5 shows average PE and PSD values by BDL approximation (trained on the full training set i.e., training plus validation). Overall, while mean PE is similar for both MC Dropout and Deep Ensembles, mean PSD for Deep Ensembles is around two-thirds (65%) higher than for MC Dropout, showing higher variability of predictions. If we exclude studies with true label disagreement, the discrepancy among the two BDL approximations further increases to around 79%.

Next, I compared TLD studies with studies where the true labels agreed (TLA). Once the (full train) data are binarised via U-Random, the percentage of TLD studies rises to around 10.8%. These studies have higher predictive/epistemic uncertainty compared to TLA studies across both BDL approximations. In particular, both MC Dropout and Deep Ensembles yield mean PEs that are around 4 times higher for TLD studies compared to TLA studies. Furthermore, while MC Dropout's mean PSD for TLD studies is almost three-quarters (73%) higher compared to TLA studies, Deep Ensembles' mean PSD for TLD studies is only around a quarter (27%) higher than for TLA studies.

To compare epistemic uncertainty with human uncertainty, I evaluated the point-biserial correlation ($R_{pb}$) between PSD and true-label disagreement (Table 6). For MC Dropout, this results in an $R_{pb}$ of around 0.35 (with a p-value significantly lower than 0.05 for this and all subsequent tests). This shows that the correlation between the two variables is statistically significant and that there is a moderate positive correlation. The $R_{pb}$ for MC Dropout is more than double the $R_{pb}$ for Deep Ensembles (around 0.16). I also evaluated CheXpert and NegBio separately, with MC Dropout showing higher correlations of around 0.54 and 0.48, respectively. In both cases, MC Dropout's correlation values are also more than double those of the Deep Ensembles, demonstrating that MC Dropout better captured TLD studies through higher epistemic uncertainty.

For predictive (total) uncertainty, I evaluated the point-biserial correlation between PE and TLD (Table 7). Here, the discrepancies between MC Dropout and Deep Ensembles are a lot subtler, as both BDL approximations yield an $R_{pb}$ of around 54%-55%. Furthermore, when evaluating CheXpert and NegBio separately, the two BDL approximations achieve around 85% for CheXpert and 77% for NegBio. Such cases show strong positive correlations, as around 72% (CheXpert) and 59%-60% (NegBio) of their variability can be explained by the relationship between PE and CheXpert/NegBio's uncertain labels. In all cases, MC Dropout performed slightly better than Deep Ensembles. In conclusion, the results for epistemic and predictive uncertainty show that MC Dropout better captured 'human' uncertainty compared to Deep Ensembles.

TABLE 5
U-RANDOM SUMMARY STATISTICS FOR PREDICTIVE ENTROPY
(PE) AND PREDICTIVE STANDARD DEVIATION (PSD)

| True Label Agreement / Disagreement | MC Dropout | | Deep Ensembles | |
|---|---|---|---|---|
| | *Mean PE* | *Mean PSD* | *Mean PE* | *Mean PSD* |
| TLA | 0.14 | 0.033 | 0.15 | 0.059 |
| TLD | 0.60 | 0.057 | 0.62 | 0.075 |
| Overall | 0.19 | 0.037 | 0.20 | 0.061 |

TABLE 6
EPISTEMIC UNCERTAINTY RESULTS (U-RANDOM)

| BDL Approximation | 'Human' Uncertainty Approach | | |
|---|---|---|---|
| | *TLD* | *Chex Uncertain* | *Neg Uncertain* |
| MC Dropout | 0.3472 | 0.5367 | 0.4751 |
| Deep Ensembles | 0.1582 | 0.2320 | 0.1917 |

TABLE 7
PREDICTIVE UNCERTAINTY RESULTS (U-RANDOM)

| BDL Approximation | 'Human' Uncertainty Approach | | |
|---|---|---|---|
| | *TLD* | *Chex Uncertain* | *Neg Uncertain* |
| MC Dropout | 0.5462 | 0.8502 | 0.7745 |
| Deep Ensembles | 0.5441 | 0.8470 | 0.7689 |

*D. Common Errors*

Cases where the model may have misjudged its uncertainty are of particular interest. I specifically analysed instances where the model was highly confident (i.e., confidence close to 0 or 1), but the 'human' interpretation was uncertain (i.e., CheXpert and NegBio disagreed). Among the resulting studies, several included extracts such as:

- *"Borderline size of the cardiac silhouette without pulmonary edema"*
- *"Unchanged borderline size of the cardiac silhouette, no pulmonary edema"*

The above excerpts clearly suggest the absence of Edema, as labelled by CheXpert and correctly predicted (with high confidence) by MC Dropout / Deep Ensembles. However, many of these studies were also labelled 'uncertain' by NegBio, which may have been triggered by the presence of the term 'borderline'. This suggests that, while CheXpert/NegBio are useful proxies for human uncertainty, their predictions should not be taken as definitive in every case.

*E. U-Ones and U-Zeros*

I repeated the previous analysis using the CheXpert-recommended Binary Mapping approaches i.e., U-Ones (replace all 'uncertain' cases with 'positive') and U-Zeros (resp. with 'negative'). Table 8 shows model performance results for U-Ones, which outperformed U-Random across both BDL approximations and the underlying BERT model. Specifically, the single-run results show around 97% accuracy and F1 score, which is confirmed by 5-fold cross-validation (around 98% for both metrics). MC Dropout and Deep Ensembles achieve around 98%-99% for both accuracy and F1 score, showing near equivalency with the CheXpert labeller. However, the results shift significantly when we evaluate uncertainty. Table 9 and Table 10 show U-Ones' results for epistemic and predictive uncertainty (respectively). Almost all correlations are much weaker compared to their U-Random counterparts, especially when evaluating CheXpert and NegBio separately (a few are even negative!). This suggests that when uncertain studies are handled via U-Random, 'human' (linguistic) uncertainty is better captured compared to U-Ones.

TABLE 8
U-ONES MODEL PERFORMANCE

| Performance Metric | Method | | | |
|---|---|---|---|---|
| | *Single Run* | *Cross Validation* | *MC Dropout* | *Deep Ensembles* |
| Accuracy | 0.9675 | 0.9818 | 0.982 | 0.9807 |
| F1 Score | 0.9744 | 0.985 | 0.9852 | 0.9841 |
| Precision | 0.9716 | | | |
| Recall | 0.9771 | | | |

TABLE 9
EPISTEMIC UNCERTAINTY RESULTS (U-ONES)

| BDL Approximation | 'Human' Uncertainty Approach | | |
|---|---|---|---|
| | *TLD* | *Chex Uncertain* | *Neg Uncertain* |
| MC Dropout | 0.1498 | 0.0250 | -0.0323 |
| Deep Ensembles | 0.1647 | 0.0121 | -0.0500 |

TABLE 10
PREDICTIVE UNCERTAINTY RESULTS (U-ONES)

| BDL Approximation | 'Human' Uncertainty Approach | | |
|---|---|---|---|
| | *TLD* | *Chex Uncertain* | *Neg Uncertain* |
| MC Dropout | 0.2072 | 0.0342 | -0.0460 |
| Deep Ensembles | 0.2005 | 0.0180 | -0.0575 |

Next, I evaluated model performance and uncertainty results for U-Zeros. While model performance is slightly inferior compared to U-Ones, U-Zeros still outperformed U-Random across both BDL approximations and the underlying BERT model, as they all achieved around 96%-97% accuracy and F1 score (Table 11). Just like U-Ones, however, U-Zeros is less effective than U-Random on uncertainty results, as all correlations are significantly weaker (Table 12 and Table 13). All in all, despite its lower model performance, U-Random yielded more promising uncertainty results compared to CheXpert's U-Ones and U-Zeros. For this reason, U-Random was chosen as the preferred uncertainty approach.

TABLE 11
U-ZEROS MODEL PERFORMANCE

| Performance Metric | Method | | | |
|---|---|---|---|---|
| | *Single Run* | *Cross Validation* | *MC Dropout* | *Deep Ensembles* |
| Accuracy | 0.9711 | 0.9688 | 0.9682 | 0.9679 |
| F1 Score | 0.9667 | 0.9617 | 0.9609 | 0.9609 |
| Precision | 0.9748 | | | |
| Recall | 0.9587 | | | |

TABLE 12
EPISTEMIC UNCERTAINTY RESULTS (U-ZEROS)

| BDL Approximation | 'Human' Uncertainty Approach | | |
|---|---|---|---|
| | *TLD* | *Chex Uncertain* | *Neg Uncertain* |
| MC Dropout | 0.0805 | 0.0246 | 0.0232 |
| Deep Ensembles | 0.0699 | 0.1000 | 0.0864 |

TABLE 13
PREDICTIVE UNCERTAINTY RESULTS (U-ZEROS)

| BDL Approximation | 'Human' Uncertainty Approach | | |
|---|---|---|---|
| | *TLD* | *Chex Uncertain* | *Neg Uncertain* |
| MC Dropout | 0.0988 | 0.0206 | 0.0197 |
| Deep Ensembles | 0.0824 | 0.1103 | 0.0963 |

*F. Test Performance*

I tested the final configuration (U-Random) on unseen data by running BERT on the full training set and evaluating it once on the test set. The results confirm the effectiveness of this setup, achieving approximately 88% test accuracy and 90% test F1 score (see Table 14). Precision was particularly high at around 97%.

So far, models have been trained and evaluated using CheXpert labels exclusively. To assess the model's robustness with out-of-distribution labels, I compared predictions from the CheXpert-trained model with NegBio labels. Although performance metrics were worse with OOD labels compared to CheXpert's in-distribution (ID) labels, the difference was only around 1%-2% across all metrics (except for recall, where OOD slightly outperformed ID). These results demonstrate the model's reliability and its ability to generalise well to OOD data.

TABLE 14
TEST PERFORMANCE ON IN-DISTRIBUTION (ID) AND OUT-OF-DISTRIBUTION (OOD) LABELS

| Performance Metric | Test Performance | |
|---|---|---|
| | *ID* | *OOD* |
| Accuracy | 0.881 | 0.8731 |
| F1 Score | 0.9018 | 0.8941 |
| Precision | 0.9694 | 0.95 |
| Recall | 0.843 | 0.8444 |

V. DISCUSSION

The results of this study provide important insights into the relationship between predictive uncertainty and human/linguistic uncertainty in the context of chest radiograph interpretation. The performance metrics of BERT on the U-Random labels (with validation accuracies and F1 scores around 88%-89%) and above all on U-Ones and U-Zeros (validation accuracies / F1 scores in the 96%-98% region) indicate a strong baseline performance for binary classification tasks in medical NLP. However, the anticipated superiority of Bayesian Deep Learning approximations, specifically MC Dropout and Deep Ensembles, was not fully realised in practice. This may be attributed to the limited number of predictions (10) generated per study due to computational constraints, which might have restricted the ability of these models to accurately quantify uncertainty. Last, but not least, while U-Random yielded good overall performance on both in-distribution and out-of-distribution data, its recall (around 84%-86% across all sets) could be improved further, as identifying patients with medical conditions is crucial in healthcare settings.

Optimising predictive performance is insufficient, as quantifying uncertainty is equally needed. Studies with true label disagreement between the CheXpert and NegBio rule-based labellers showed higher predictive/epistemic uncertainty (as measured by predictive entropy and predictive standard deviation) compared to studies where the two labellers agreed. Despite this, U-Random's point-biserial correlations between model and linguistic uncertainty, although positive, were not consistently high. While the presence/absence of uncertainty labels in CheXpert and NegBio resulted in correlations in the 77%-85% region for predictive uncertainty, the remaining correlations showed at best moderate associations. Nonetheless, U-Random performed better than U-Ones/U-Zeros, whose (point-biserial) correlations were practically non-existent. Overall, while models captured some aspects of human uncertainty, they still fell short of fully representing the complexity and variability inherent in human decision-making for medical applications.

The variability in label distribution across training, validation, and test sets further complicated the interpretation of results, highlighting the importance of dataset balance and the potential need for stratified sampling in future studies. The higher prevalence of positive cases in the test set, coupled with the discrepancies in uncertainty labels between CheXpert and NegBio, underscores the challenges of using rule-based labellers as ground truth in the absence of human annotations. This study's reliance on U-Random binarisation as a middle-ground approach between U-Ones and U-Zeros provided a valuable perspective, although it also introduced additional variability that may have impacted model performance.

All in all, the findings suggest that while current methods for uncertainty quantification are a step in the right direction, there is still significant room for improvement. By advancing our understanding of how models can better align with human uncertainty, we can move closer to developing AI systems that are not only accurate but also trustworthy and reliable in high-stakes clinical environments.

## VI. FURTHER ANALYSIS

Overall results, both in terms of modelling and uncertainty, may be improved with more/better data. For example, additional chest X-ray datasets, such as CheXpert and ChestX-ray14, could be explored. To improve the data from this study (MIMIC-CXR-JPG), stratified sampling may be employed, as it would achieve more balanced splits between classes. Furthermore, the preprocessing of text reports could be refined further, for example by replacing common medical abbreviations (e.g. CHF, SVC) with their spelled-out equivalents (Congenital Heart Failure, Superior Vena Cava).

Model performance, and especially U-Random's recall, may be increased by exploring alternative transformer-based models such as ClinicalBERT and RoBERTa. We could even move beyond transformers by evaluating RNN/CNN architectures. It is also worth noting that hyperparameters for MC Dropout, Deep Ensembles, and the underlying BERT were found empirically and, for this reason, a more systematic search (e.g. Grid/Random Search) may be beneficial (albeit computationally expensive). Even just lowering the decision threshold (e.g., 0.4 instead of the current 0.5) could potentially increase recall (but also decrease precision). Plotting a Precision-Recall curve (or an ROC curve) would be useful to see how using different thresholds affects both performance metrics.

Due to computational constraints, MC Dropout and Deep Ensembles were both used to generate only 10 predictions per study. Increasing the number of predictions may improve uncertainty (but also model performance) results. Alternative uncertainty-aware models, such as Bayesian Neural Networks and Gaussian Processes, could also be employed. Last, but not least, exploring alternative uncertainty approaches, such as Self-Training and 3-Class Classification, could be useful to boost uncertainty results. As performance is improved, the analysis on Edema may be expanded to other chest observations, such as Cardiomegaly and Pleural Effusion.

Ideally, we would like to have multiple human labels for chest X-ray data instead of relying on rule-based labellers to approximate human uncertainty. With multiple human labels available, Referral Learning for chest radiograph interpretation may be explored. A machine-human collaboration has the potential to outperform both the model and the human working independently, with improved workflow prioritisation. Furthermore, having machines take on high-certainty cases may free up time and resources for doctors to spend on more complex cases. This would be especially beneficial in resource-poor countries, where radiology services are scarce, but even developed countries may benefit from Referral Learning with large-scale screening and nationwide health initiatives.

## VII. CONCLUSION

This study explored the intersection of linguistic and predictive uncertainty in the context of chest radiograph interpretation using Bayesian Deep Learning approximations. While BERT, enhanced with Monte Carlo Dropout and Deep Ensembles, achieved commendable predictive performance, the modest correlation with linguistic uncertainty reveals the limitations of current methods in fully capturing the nuances of human decision-making. The findings underscore the need for further research into more advanced uncertainty quantification techniques and highlight the potential of referral learning approaches, where models and human experts collaborate to optimise clinical outcomes. Future work should focus on refining uncertainty estimation methods, exploring alternative labelling strategies, and incorporating human-in-the-loop approaches to better bridge the gap between machine predictions and human expertise for medical applications. Advancing our understanding of how models can better align with human uncertainty brings us closer to developing AI systems that are not only accurate but also trustworthy and reliable in high-stakes clinical settings.

### ETHICS AND CONFLICT OF INTEREST